\documentclass[a4paper,fleqn]{cas-sc}

\usepackage[authoryear,longnamesfirst]{natbib}
\usepackage{float}
\usepackage{algorithm}
\usepackage{algpseudocode}
\usepackage{placeins}
\usepackage{xcolor}

\begin{document}
\let\WriteBookmarks\relax
\def\floatpagepagefraction{1}
\def\textpagefraction{.001}
\shorttitle{\textbf{GAIA}: \textbf{G}eometry-\textbf{A}ware \textbf{I}nfrastructure-\textbf{A}nchored Denoiser for UWB Sensing and Work-Zone Reconstruction}

\shortauthors{Tang et~al.}

\title [mode = title]{\textbf{GAIA}: \textbf{G}eometry-\textbf{A}ware \textbf{I}nfrastructure-\textbf{A}nchored Denoiser for UWB Sensing and Work-Zone Reconstruction}

\author[1]{Weizhe Tang}

\affiliation[1]{organization={Department of Civil \& Environmental Engineering, University of Wisconsin-Madison},
    city={Madison},
    postcode={53706},
    state={Wisconsin},
    country={United States}}

\author[1]{Jiaxi Liu}[orcid=0009-0001-2749-6435]
\cormark[1]
\ead{jliu2487@wisc.edu}

\author[1]{Junwei You}

\author[1]{Steven T. Parker}

\author[2]{Pei Li}
\affiliation[2]{organization={Department of Civil and Architectural Engineering and Construction Management, University of Wyoming},
    city={Laramie},
    postcode={82071},
    state={Wyoming},
    country={United States}}

\author[1]{Sikai Chen}

\author[3]{Meng Ran}

\affiliation[3]{%
    organization={School of Computer Science and Technology, Chongqing University of Posts and Telecommunications},
    city={Chongqing},
    postcode={400065},
    country={China}
}

\author[1]{Bin Ran}

\cortext[cor1]{Corresponding author}

\begin{abstract}
Accurate perception of work-zone geometry is critical for intelligent transportation systems, and ultra-wideband (UWB) sensing offers a low-cost path toward infrastructure-aided work-zone reconstruction. However, UWB ranging in outdoor work zones is often affected by non-line-of-sight (NLOS) propagation, burst noise, and long-tail errors, which can severely distort downstream spatial reconstruction. Existing approaches mainly address signal-level range denoising and do not explicitly model the geometric structure needed for reliable boundary reconstruction.

In this work, we present GAIA, a geometry-aware, infrastructure-anchored learning framework for UWB denoising and work-zone reconstruction. GAIA estimates geometry-consistent UWB distances by coupling temporal range modeling with latent anchor-layout estimation and deterministic distance projection. The framework keeps range denoising as the supervised prediction task while orienting the learned distances toward boundary-consistent work-zone reconstruction.

We evaluate GAIA primarily on a real-world outdoor UWB dataset with synchronized UWB, GNSS, and IMU measurements under LOS and NLOS conditions. To complement this evaluation, we use a real-data-calibrated stress-test simulator to examine robustness under stronger NLOS corruption and long-tail ranging errors. On the real-world dataset, GAIA achieves the lowest overall range MSE and the highest polygon IoU among the evaluated filtering-based and learning-based baselines, reducing overall MSE by 18.4\% and improving polygon IoU by 15.5\% over PoseMLP. Supplementary stress-test simulation and ablation results further show that the geometry-aware components improve boundary-level reconstruction under severe ranging noise. These results indicate that geometry-aware range denoising is a promising direction for infrastructure-aided work-zone reconstruction, and that GAIA provides a boundary-oriented UWB reconstruction framework that links range correction to spatially coherent work-zone geometry estimation.
\end{abstract}

% \includegraphics{figs/cas-grabs.pdf}

\begin{highlights}
\item \textbf{Geometry-aware work-zone reconstruction:} UWB mapping is formulated as boundary reconstruction, with range denoising serving the geometric objective.
\item \textbf{Latent geometry-guided denoising:} A latent anchor layout is inferred and fed back into denoising as an explicit spatial prior.
\item \textbf{Real-data-centered validation with stress testing:} GAIA is evaluated primarily on real outdoor UWB measurements, with a real-data-calibrated simulator used as a supplementary stress-test environment for severe NLOS and long-tail errors.
\end{highlights}

\maketitle
\section{Introduction}

Work zones are among the most hazardous and geometrically dynamic environments on public roads. In 2023 alone, work-zone crashes caused 898 fatalities and more than 40{,}000 injuries in the U.S. \cite{nsc2024workzone_injuryfacts}. Work zones introduce rapidly changing drivable boundaries through lane narrowing, lateral shifts, tapers, and temporary barriers, creating geometric conditions that differ substantially from normal roadway layouts \cite{dehman2021workzone_cav_review}. These layouts are often irregular and frequently revised during construction, making drivable boundaries difficult to represent and track consistently.
Providing an accurate and frequently updated estimate of the spatial boundary of the active
work zone is therefore a foundational requirement for compliant routing, hazard buffering,
and safe passage \cite{usdot2024wzdx}.

For work-zone sensing, the central question is where the traversable boundary lies. We therefore formulate work-zone perception as a geometry reconstruction task and emphasize boundary-level metrics such as intersection-over-union~(IoU) and Hausdorff distance. Existing approaches to work-zone awareness predominantly frame the problem as detection, focusing on traffic control devices, workers, or equipment. Survey-grade or LiDAR-based mobile mapping systems can recover detailed roadway geometry, but they require dedicated calibrated equipment and cannot continuously follow changing conditions \cite{habib2018lidar_lane_width_workzones}. Vision-based pipelines detect temporary traffic control devices~(TTCDs) and infer scene extent through topology reasoning \cite{seo2022ttcd_crc,zuo2023urban_workzone_detection_sizing}, but they remain sensitive to occlusion, adverse lighting, and limited annotated work-zone data. Trajectory-based crowdsourcing methods infer drivable regions from vehicle paths \cite{chen2023crowdsourcing_workzone_mapping}, yet they can lag behind abrupt layout changes and degrade under sparse observations. More broadly, the HD-map update literature shows that maintaining accurate geometric representations of dynamic road environments remains difficult and expensive \cite{li2022hdmapnet,liao2022maptr}. Despite their differences, these approaches share a common limitation: they optimize for element-level detection or localization while treating the drivable work-zone boundary as an implicit by-product. This indirect formulation obscures the geometric structure of the reconstruction problem and limits reasoning about boundary-level errors. It also raises the question of which sensing modality can provide reliable, deployable geometric constraints for frequently updated boundary reconstruction in dynamic work zones.

Ultra-wideband~(UWB) radio ranging offers a compelling low-cost complement
to onboard perception:
fine time-of-flight resolution enables accurate distance measurement
\cite{gezici2005uwb_localization_spm,alarifi2016uwb_indoor_review},
hardware is commercially available and field-deployable at modest cost
\cite{volpi2023lowcost_uwb_rtls},
and UWB-based real-time locating systems~(RTLS) have already demonstrated utility in
construction-site worker and equipment tracking
\cite{maalek2016uwb_construction_accuracy,ochoa2024uwb_roadworker_safety}.
In the V2I (vehicle-to-infrastructure) setting, UWB-enabled roadside units~(UWB-RSUs)
placed along the work-zone perimeter can provide multi-anchor geometric constraints to a
vehicle-mounted tag as the vehicle traverses the zone, yielding a direct, infrastructure-side
signal path to the work-zone boundary. The resulting multi-anchor measurements provide structured geometric information for boundary reconstruction as coupled range observations. However, UWB ranging is highly sensitive to NLOS propagation, occlusion, and environmental interference, which can introduce significant bias and instability in real-world deployments \cite{wang2023uwb_nlos_survey}.

\begin{figure}[]
    \centering
    \includegraphics[width=0.8\linewidth]{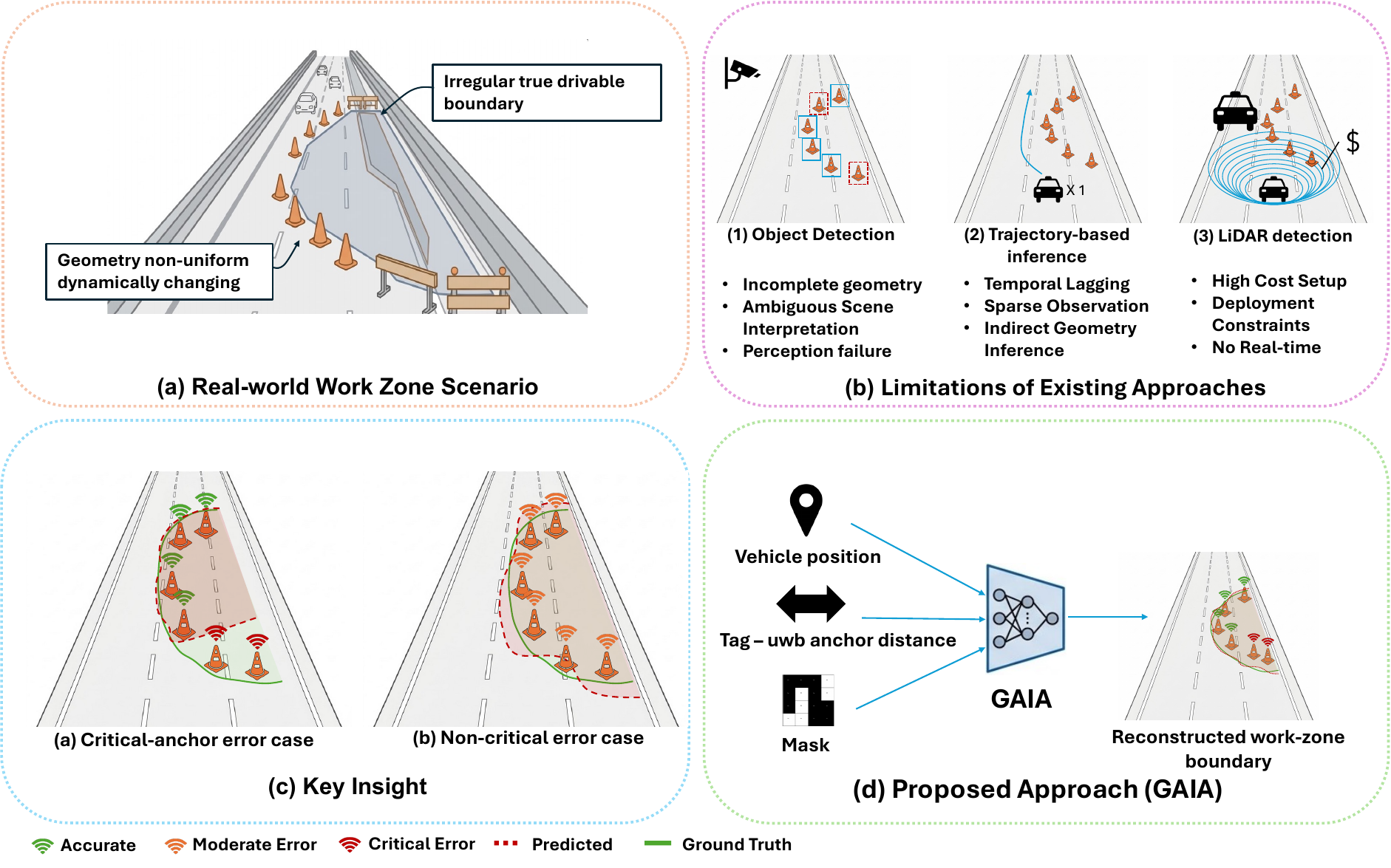}
    \caption{Overview of the proposed geometry-first UWB denoising framework and motivating examples.}
    \label{fig:overview}
\end{figure}

To address these challenges, recent work has explored learning-based correction of UWB ranges, showing that learned denoising can suppress NLOS-induced bias \cite{wang2023uwb_nlos_survey,angarano2021robust_uwb_deep_edge} and improve downstream ranging accuracy under work-zone propagation conditions. However, these approaches primarily formulate the problem as range denoising: the objective is to recover accurate anchor-wise distances, while work-zone geometry reconstruction remains a secondary downstream outcome \cite{liu2026v2iworkzonegeometry}. This formulation leaves an important gap. In dynamic work-zone environments, UWB measurements are frequently affected by NLOS propagation, occlusion, and environmental variability, making it difficult to maintain reliable measurements across all anchors. More importantly, boundary reconstruction is a geometric problem in which anchor errors contribute unevenly to the final reconstructed shape. As illustrated in Fig.~\ref{fig:overview}(c), reconstruction quality is often dominated by geometrically critical anchors. In Case A, most anchors remain accurate, yet a small number of critical anchor errors creates a large missing region in the reconstructed boundary. In contrast, Case B contains moderate errors across all anchors, producing noisier local geometry while preserving the overall boundary structure. This example shows why reliable work-zone reconstruction requires more than average ranging error minimization: reconstruction quality depends heavily on spatially critical errors. Moreover, real outdoor work-zone UWB data are costly to collect, and measured datasets may not cover severe NLOS, burst-noise, and long-tail error cases. This motivates a real-data-centered evaluation strategy complemented by controlled stress-test simulations, with real data serving as the primary evidence.

To address these limitations, we present \textbf{GAIA} (\underline{G}eometry-\underline{A}ware \underline{I}nfrastructure-\underline{A}nchored
denoiser), a learned UWB range-denoising framework designed for work-zone geometry reconstruction. GAIA uses multi-anchor UWB observations and geometry-aware regularization to make denoised ranges more consistent with downstream boundary reconstruction. As demonstrated in the experiments, GAIA outperforms the evaluated filtering-based and learning-based baselines on polygon IoU while remaining competitive on range error. Our contributions are as follows:

\begin{itemize}
\item \textbf{Geometry-aware formulation for work-zone reconstruction.} We formulate UWB work-zone mapping as boundary-oriented range denoising, where supervised range prediction is regularized by latent layout and geometry-consistency objectives to support downstream reconstruction.

\item \textbf{Geometry-aware latent layout reconstruction.}
We design a geometry-aware denoising framework that first infers a latent anchor layout from temporal multi-anchor UWB observations and then feeds the inferred geometry back into the denoising process as an explicit spatial prior. This enables the model to jointly enforce temporal consistency and spatially coherent boundary reconstruction from coupled multi-anchor ranging signals.

\item \textbf{Real-world validation with supplementary stress testing.}
We validate GAIA primarily on a real outdoor UWB dataset with synchronized UWB, GNSS, and IMU measurements under LOS and NLOS conditions. On the real-world dataset, GAIA achieves an 18.4\% reduction in overall range MSE and a 15.5\% increase in polygon IoU over PoseMLP. We further use a real-data-calibrated stress-test simulator to evaluate model behavior under stronger NLOS corruption, long-tail ranging errors, and controlled geometry variation.
\end{itemize}

\section{Related Work}

\subsection{Work-Zone Perception: Detection, Mapping, and Geometry Estimation}
\label{sec:rw_workzone}

Existing work on work-zone perception mainly addresses detection, mapping, and localization for autonomous driving systems. Vision-based approaches detect temporary traffic control devices and infer work-zone extent from scene structure \cite{seo2022ttcd_crc,zuo2023urban_workzone_detection_sizing}, while LiDAR-based mobile mapping systems recover detailed roadway geometry using calibrated survey equipment \cite{habib2018lidar_lane_width_workzones}. Crowdsourcing approaches estimate drivable regions from accumulated vehicle trajectories \cite{chen2023crowdsourcing_workzone_mapping}, and recent learning-based HD-map frameworks generate vectorized lane representations directly from sensor observations \cite{li2022hdmapnet,liao2022maptr,liu2023intersectionmap_roadside_lidar}. The ROADWork benchmark further shows that modern foundation models still struggle in work-zone environments and identifies drivable path estimation as an open problem \cite{ghosh2025roadwork_iccv}.

Despite their differences, these approaches primarily optimize for object detection, localization, or map generation, while treating the traversable work-zone boundary as an implicit by-product. Planning and safety, however, depend directly on the drivable boundary itself. Vision systems remain sensitive to occlusion and lighting conditions, while LiDAR mapping requires dedicated infrastructure and does not naturally support continuous adaptation to rapidly changing layouts. These limitations motivate UWB roadside sensing as a deployable geometric sensing modality for work-zone boundary reconstruction.

\subsection{UWB Ranging: Capabilities, NLOS Challenges, and Infrastructure Deployments}

Ultra-wideband~(UWB) ranging enables centimeter-scale distance estimation through precise time-of-flight measurement and has been widely adopted in real-time locating systems~(RTLS) \cite{gezici2005uwb_localization_spm,alarifi2016uwb_indoor_review}. Modern IEEE~802.15.4a/z UWB transceivers support protocols such as two-way ranging~(TWR) and time-difference-of-arrival~(TDoA), and can be deployed as low-cost field-operable sensing nodes \cite{volpi2023lowcost_uwb_rtls}. Recent studies have demonstrated UWB deployment in industrial and construction environments, including work-zone safety systems where anchors are mounted directly on traffic cones \cite{maalek2016uwb_construction_accuracy,ochoa2024uwb_roadworker_safety}. These properties make UWB an attractive sensing modality for vehicle-to-infrastructure work-zone perception.

Outdoor UWB ranging, however, remains sensitive to non-line-of-sight propagation, multipath reflections, and dynamic occlusion, which introduce systematic ranging bias \cite{wang2023uwb_nlos_survey}. Existing V2I UWB systems mainly use anchor measurements for localization, geofencing, or proximity warning, while treating measurements from different anchors independently \cite{liu2026v2iworkzonegeometry}. As a result, the geometric structure formed by multi-anchor layouts has not been fully exploited for work-zone boundary reconstruction, motivating learning-based ranging correction methods that incorporate geometric consistency.

\subsection{Learning-Based Range Denoising and Geometry-Aware Reconstruction}
\label{sec:rw_learning}

Learning-based methods have substantially improved UWB ranging under non-line-of-sight conditions. Early studies applied machine learning to waveform and channel features for ranging-error mitigation and LOS/NLOS identification \cite{wymeersch2012machine,jiang2020uwb_nlos_dl_commlett,cui2021uwb_nlos_cnn_morlet}. Subsequent work introduced deeper architectures, including autoencoder-based ranging correction and lightweight edge-deployable regression networks for work-zone environments \cite{fontaine2020uwb_autoencoder_access,angarano2021robust_uwb_deep_edge}. More recent approaches adopt attention and transformer architectures to jointly model channel characteristics and multi-anchor observations for ranging refinement and position correction \cite{pei2024fcn_attention_uwb,yang2025fuzzy_bert_uwb_tim,coppens2025transformer_tdoa_uwb}. In parallel, several studies have explored structural or location-aware learning to incorporate spatial relationships among anchors \cite{coene2024location_aware_uwb_sensors,luo2022geometric_dl_ipin}.

Despite these advances, existing methods remain primarily optimized for anchor-wise ranging accuracy. Their objectives are typically defined by mean range error or LOS/NLOS classification accuracy, without directly considering the downstream geometry reconstruction task. Work-zone boundary reconstruction is inherently geometry-sensitive: errors on geometrically critical anchors can severely distort the reconstructed boundary, whereas similar errors on non-critical anchors may have limited impact. Consequently, minimizing average ranging error alone is insufficient for reliable boundary reconstruction. In addition, many existing methods rely heavily on simulated or semi-controlled datasets, raising concerns about deployment realism and transfer to outdoor work-zone environments.

\subsection{Real-World Evaluation and Simulation-Based Stress Testing for UWB Systems}

Real-world UWB data are essential for evaluating deployment-oriented ranging and reconstruction methods because outdoor UWB errors depend on propagation conditions, anchor placement, obstruction patterns, and vehicle motion \cite{wang2021semi_supervised_uwb_wcl,li2023semi_supervised_uwb_waveform}. At the same time, real-world data collection is expensive and may not fully cover severe NLOS, burst noise, and long-tail ranging errors. Prior V2I work-zone UWB studies therefore combine measured field data with controlled simulation to support robustness evaluation and ablation analysis \cite{liu2026v2iworkzonegeometry}. Following this perspective, our evaluation is centered on real-world UWB measurements, with simulation used as a supplementary stress-test environment.

Motivated by the limitations discussed above, we propose GAIA, a geometry-aware UWB denoising framework for boundary-consistent work-zone reconstruction, which is described in Sec.~\ref{sec:Methodology}.

\section{Methodology}
\label{sec:Methodology}

\subsection{Method Overview}

Our goal is geometry-aware UWB range denoising for work-zone boundary reconstruction.
Given a vehicle traversing a work zone, we observe its location sequence together with multi-anchor UWB ranging measurements.
Specifically, the input consists of a vehicle location sequence
$p_{\mathrm{seq}} \in \mathbb{R}^{B \times T \times 2}$,
where each element contains the 2D vehicle coordinates over time,
the raw UWB distance measurements
$d_{\mathrm{seq}}^{\mathrm{obs}} \in \mathbb{R}^{B \times T \times N}$ from the vehicle to cones equipped with UWB anchors,
and a padding mask
$m_{\mathrm{seq}} \in \{0,1\}^{B \times T \times N}$ indicating valid observations.
Here, $B$ denotes the batch size, $T$ denotes the temporal length, and $N$ denotes the number of anchors.

We assume that the vehicle pose sequence $p_{\mathrm{seq}}$ is provided by an onboard positioning module, such as GNSS/INS or SLAM, and we focus on denoising UWB ranges for latent anchor-layout and work-zone boundary reconstruction. In the current experiments, we use the RTK-GNSS trajectories available in the dataset as pose input to isolate the effect of UWB range denoising on geometry reconstruction. This corresponds to an upper-bound setting for pose-conditioned reconstruction; characterizing GAIA's sensitivity to noisier pose sources is left for future work. The vehicle pose is therefore treated as a known input. GAIA localizes anchor positions and reconstructs the work-zone boundary shape, and the evaluation accordingly focuses on range accuracy, anchor position error, polygon IoU, and Hausdorff distance. Vehicle or tag localization lies outside the scope of this study.

\begin{figure}[]
    \centering
    \includegraphics[width=\linewidth]{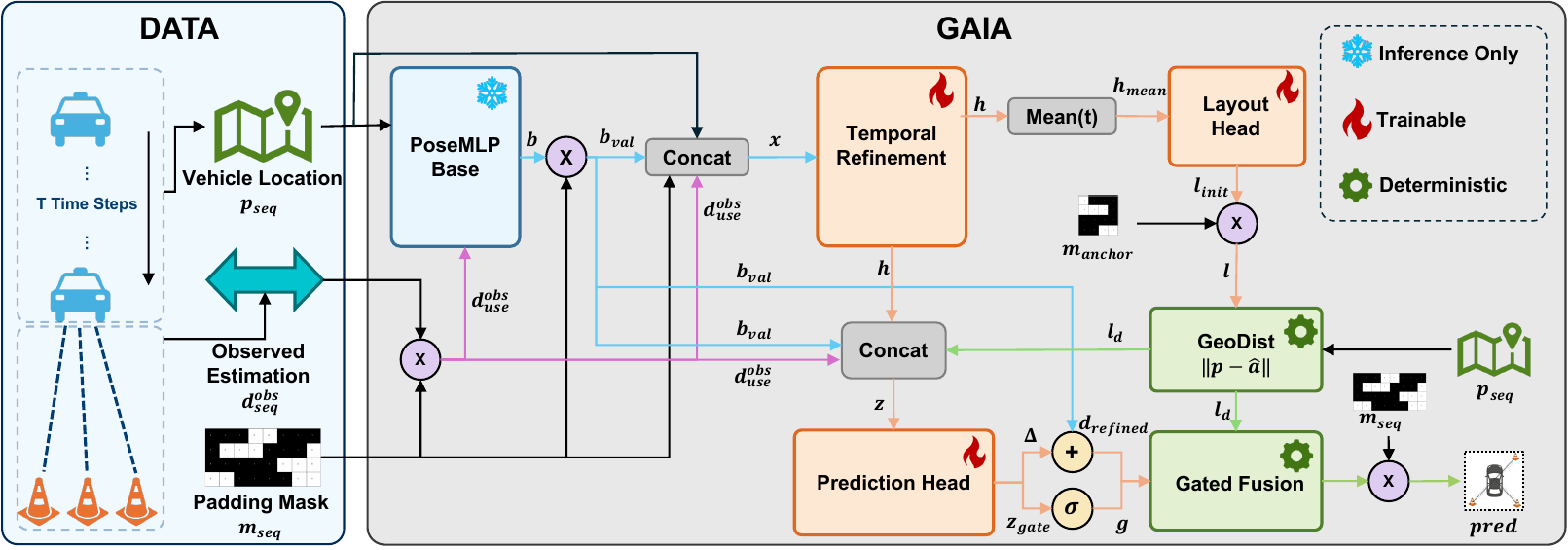}
\caption{Overview of the proposed geometry-aware UWB denoising framework (GAIA). The model takes vehicle trajectories, multi-anchor UWB ranges, and validity masks as input, and produces denoised distances through the PoseMLP Base, Temporal Refinement, Layout Head, GeoDist, Prediction Head, and Gated Fusion modules. The PoseMLP Base module is pretrained and frozen, providing a stable per-step initialization. The diagram distinguishes pretrained frozen components, trainable neural modules, and deterministic geometric operators. The Layout Head predicts a latent anchor configuration, from which geometry-consistent distances are computed by GeoDist and combined with learned residual corrections through gated fusion.}
    \label{fig:framework}
\end{figure}

The key idea of GAIA is to infer a latent geometric representation from temporal multi-anchor UWB observations and feed the inferred geometry back into the denoising process. As illustrated in Fig.~\ref{fig:framework}, the model contains six modules:
a PoseMLP Base module, a Temporal Refinement module, a Layout Head module, a GeoDist module, a Prediction Head module, and a Gated Fusion module.

We process the input sequence through the modules in order. We first apply the padding mask $m_{\mathrm{seq}}$ to the raw UWB measurements $d_{\mathrm{seq}}^{\mathrm{obs}}$ to obtain valid observations, ensuring that only available anchor measurements contribute to subsequent computation. The masked observations and vehicle location sequence $p_{\mathrm{seq}}$ are then fed into a pretrained PoseMLP Base module. We keep this module frozen during training because it already provides a stable representation of per-anchor distance patterns and serves as a strong initialization for further refinement.

The resulting features are combined with the original masked observations, mask signals, and vehicle locations, and then passed into the Temporal Refinement module. This module captures temporal dependencies across the sequence and produces a latent representation that encodes temporal dynamics and cross-anchor correlations, which are critical for handling burst noise and inconsistent measurements.

From the refined temporal representation, we compute a global sequence summary and feed it into the Layout Head module to predict a latent spatial configuration of anchors. This step explicitly introduces geometric structure, allowing the model to reason about anchor arrangement as a coupled spatial configuration.

Given the predicted layout and the vehicle trajectory, the GeoDist module deterministically computes geometry-consistent distances between the vehicle and anchors. This provides a physically grounded reference signal implied by the inferred spatial configuration.

Finally, the Prediction Head module produces a correction term and a gating signal, which the Gated Fusion module uses to combine refined estimates with geometry-derived distances. This fusion allows the model to adaptively balance learned corrections and geometric consistency, yielding the final denoised distance predictions.

\subsection{Architecture of GAIA}
\label{sec:Architecture}

GAIA consists of six modules: the PoseMLP Base module, the Temporal Refinement module, the Layout Head module, the GeoDist module, the Prediction Head module, and the Gated Fusion module. We describe these modules in the order of the forward pass.

\paragraph{Masked observation construction.}
We first apply the validity mask $m_{\mathrm{seq}}$ to the raw UWB measurements $d_{\mathrm{seq}}^{\mathrm{obs}}$ to remove invalid or padded observations, obtaining the masked observation tensor $d_{\mathrm{use}}^{\mathrm{obs}} \in \mathbb{R}^{B \times T \times N}$. Here, $B$ denotes the batch size, $T$ the temporal sequence length, and $N$ the number of anchors. This masking step preserves valid anchor measurements for subsequent processing and prevents padded or invalid anchors from contributing to feature extraction or final prediction.

\paragraph{PoseMLP base module.}

The PoseMLP Base module is a pretrained pose-aware denoising network \cite{liu2026v2iworkzonegeometry} that serves as a stable backbone for spatial ranging patterns related to the underlying anchor layout. It is pretrained as a per-step denoising baseline and kept frozen in GAIA. For each time step, the module takes the masked anchor measurements together with the corresponding vehicle location and predicts an initial per-anchor distance estimate. Specifically, the per-step masked ranging vector and the 2D vehicle location are fed into a three-layer multilayer perceptron with ReLU activations. The output is an initial base estimate for all anchors at that time step, denoted as $b \in \mathbb{R}^{B \times T \times N}$. We then apply the validity mask $m_{\mathrm{seq}}$ to obtain the valid-anchor base estimate, denoted as $b_{\mathrm{val}} \in \mathbb{R}^{B \times T \times N}$.

By providing an initial estimate that already reflects coarse anchor-layout information, the PoseMLP Base module gives a strong initialization for subsequent temporal refinement and geometry-aware processing. The ablation study in Sec.~\ref{sec:Ablation} examines the effect of this frozen backbone on boundary-level reconstruction.

\paragraph{Temporal Refinement module.}
After obtaining the base estimate, we concatenate the masked observation $d_{\mathrm{use}}^{\mathrm{obs}}$, the base prediction $b_{\mathrm{val}}$, the validity mask $m_{\mathrm{seq}}$, and the vehicle location sequence $p_{\mathrm{seq}}$ along the feature dimension, forming a fused sequential feature $x \in \mathbb{R}^{B \times T \times 3N+2}$. This sequence is processed by the Temporal Refinement module, which consists of a bidirectional GRU followed by a LayerNorm layer. The bidirectional GRU captures temporal dependencies in both forward and backward directions, allowing the model to use contextual information from neighboring time steps. This is important in UWB denoising because NLOS corruption and ranging noise are often temporally correlated across frames. Because the bidirectional design uses both past and future context within each window, GAIA operates over short temporal windows and is most suitable for post-pass reconstruction or near-real-time mapping with bounded look-ahead delay; a causal temporal module is left for future work. The subsequent normalization stabilizes the hidden representation and improves optimization. The output is a latent temporal feature $h \in \mathbb{R}^{B \times T \times 2H_{\mathrm{seq}}}$, where $H_{\mathrm{seq}}$ denotes the GRU hidden dimension and the factor of two comes from concatenating the forward and backward hidden representations. This representation summarizes temporal ranging dynamics, vehicle-motion context, and cross-anchor geometric interactions, providing the basis for geometry-aware layout inference.

\paragraph{Layout Head module.}

\begin{figure}[]
    \centering
    \includegraphics[width=0.9\linewidth]{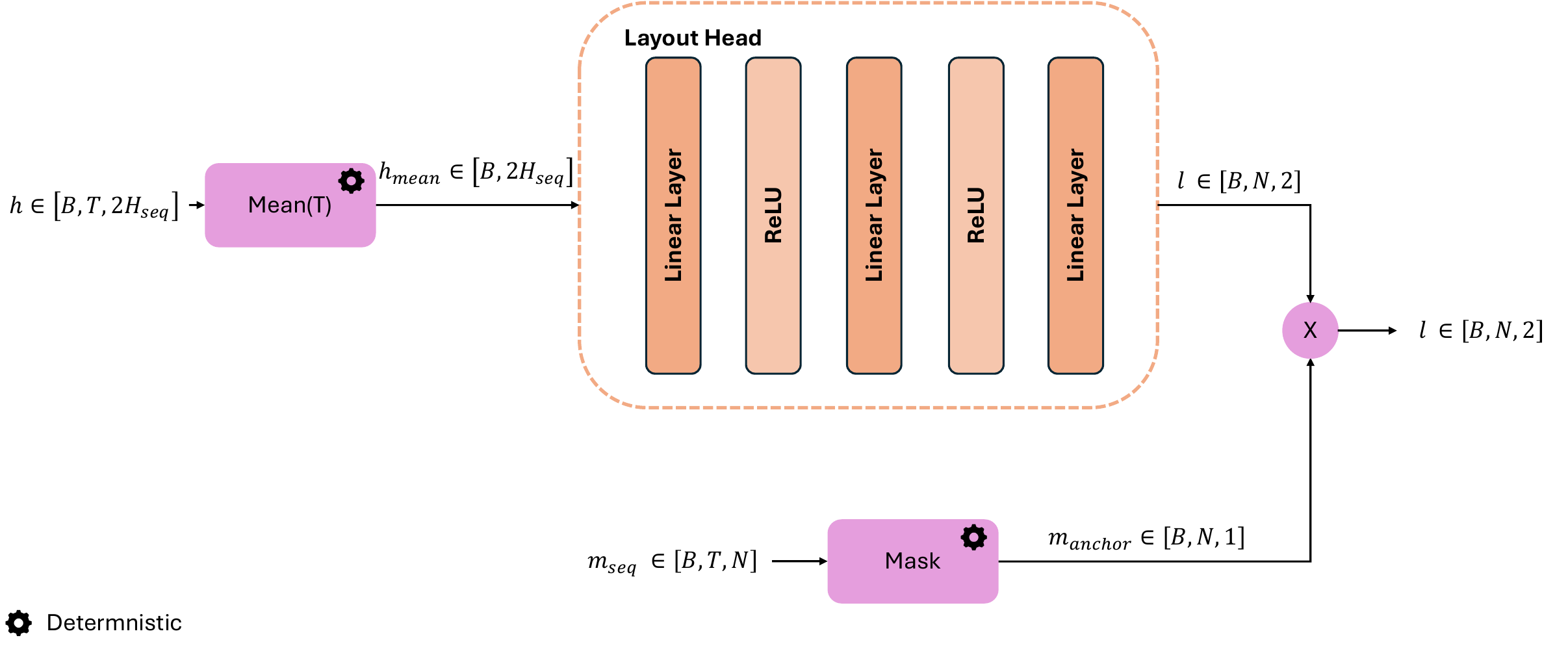}
    \caption{Detailed structure of the Layout Head module. 
The temporal feature $h \in \mathbb{R}^{B \times T \times 2H_{\mathrm{seq}}}$ is first aggregated along the time dimension using a deterministic mean operator, producing a global representation $h_{\mathrm{mean}} \in \mathbb{R}^{B \times 2H_{\mathrm{seq}}}$.
This representation is then processed by a three-layer multilayer perceptron to predict the latent anchor layout $l \in \mathbb{R}^{B \times N \times 2}$.
In parallel, an anchor-level validity mask is derived from $m_{\mathrm{seq}}$ and applied to the predicted layout via element-wise multiplication, so that only valid anchors contribute to the inferred geometry.
Deterministic operators without learnable parameters and the trainable neural layers inside the Layout Head are visually distinguished in the diagram.}
    \label{fig:layout_head}
\end{figure}

The detailed structure of the Layout Head module is illustrated in Fig.~\ref{fig:layout_head}. The temporal feature $h \in \mathbb{R}^{B \times T \times 2H_{\mathrm{seq}}}$ is first aggregated along the time dimension using a mean operator, producing a compact global representation $h_{\mathrm{mean}} \in \mathbb{R}^{B \times 2H_{\mathrm{seq}}}$. This aggregated feature is fed into a three-layer multilayer perceptron with ReLU activations, which maps the temporal representation into an initial latent anchor layout $l_{\mathrm{init}} \in \mathbb{R}^{B \times N \times 2}$.

In parallel, as shown in the lower branch of the figure, an anchor-level validity mask $m_{\mathrm{anchor}} \in \mathbb{R}^{B \times N \times 1}$ is constructed from the sequence mask $m_{\mathrm{seq}}$ and applied to the initial layout $l_{\mathrm{init}}$ through element-wise multiplication to produce the final latent layout $l \in \mathbb{R}^{B \times N \times 2}$.
This masking operation prevents invalid or padded anchors from contributing to the inferred spatial configuration, thereby maintaining consistency between the latent layout and the effective anchor set.

Beyond its architectural role, the Layout Head enables geometry-aware modeling. While earlier stages operate primarily on temporal UWB signals, this module introduces a spatial representation by inferring anchor positions from the temporal feature. This transition from sequence modeling to spatial reasoning is essential because accurate work-zone geometry reconstruction depends on the underlying anchor layout.

The inferred layout serves as the basis for subsequent geometry-aware distance computation and fusion, encouraging consistency between predicted distances and the underlying spatial structure.

As described in Secs.~\ref{sec:training} and \ref{sec:loss}, the predicted layout is further supervised by geometry-aware losses that encourage physically meaningful and globally consistent anchor configurations.

The Layout Head establishes a latent geometric representation that bridges temporal signal processing and explicit spatial reasoning, forming the foundation of GAIA's geometry-aware design. The ablation study in Sec.~\ref{sec:Ablation} examines its contribution to boundary-level reconstruction.

\par

\paragraph{GeoDist module.}
Given the predicted anchor layout $l$ and the vehicle location sequence $p_{\mathrm{seq}}$, the GeoDist module computes a geometry-consistent distance tensor as the Euclidean distance between the vehicle and each predicted anchor:
\begin{equation}
l_{d,b,t,i} = \left\| p_{b,t} - l_{b,i} \right\|_2,
\quad
l_d \in \mathbb{R}^{B \times T \times N},
\end{equation}

% \begin{equation}
% l_{d} = \left\| p_{b,t} - l_{b,i} \right\|_2,
% \quad
% l_d \in \mathbb{R}^{B \times T \times N},
% \end{equation}

where the vehicle position $p_{b,t} \in \mathbb{R}^2$ is broadcast over the $N$ anchors and the layout $l_{b,i} \in \mathbb{R}^2$ over the $T$ time steps.
This module contains no learnable parameters and is fully deterministic. Its primary role is to project the inferred anchor layout back into the ranging space, producing geometry-consistent distance estimates constrained by the underlying spatial configuration.

This geometric projection introduces an explicit spatial constraint into the denoising process, encouraging predicted distances to remain consistent with the inferred anchor layout and reducing reliance on learned temporal correlations alone.

The resulting layout distance $l_d$ serves as a key intermediate representation that connects multiple components of the framework. It is used in the Prediction Head and Gated Fusion modules to balance learned corrections with geometry-consistent estimates, and it also participates in the geometry-aware supervision described in subsequent sections. Through this design, the GeoDist module helps align the latent layout, predicted distances, and underlying physical geometry. The ablation study in Sec.~\ref{sec:Ablation} evaluates the contribution of this explicit geometric projection.

\paragraph{Prediction Head module.}
The Prediction Head module takes the temporal feature $h$, the base prediction $b_{\mathrm{val}}$, the masked observation $d_{\mathrm{use}}^{\mathrm{obs}}$, and the geometry-derived distance $l_d$, and concatenates them along the feature dimension. The resulting fused feature $z \in \mathbb{R}^{B \times T \times (2H_{\mathrm{seq}}+3N)}$ is then passed through another three-layer MLP with ReLU activations. This module outputs two quantities: a residual correction term and a gating logit. After splitting the output along the last dimension, we obtain $\Delta \in \mathbb{R}^{B \times T \times N}$
and $z_{\mathrm{gate}} \in \mathbb{R}^{B \times T \times N}.$

The residual term $\Delta$ refines the frozen PoseMLP prediction, while the gating branch determines how much the model should rely on the refined estimate versus the geometry-based estimate. This design makes the final prediction both data-adaptive and geometry-aware.

We then obtain the gating signal $g \in \mathbb{R}^{B \times T \times N}$ by applying a sigmoid activation to $z_{\mathrm{gate}}$, and compute the refined base estimate $d_{\mathrm{refined}} \in \mathbb{R}^{B \times T \times N}$ by adding the residual correction to the base prediction followed by a non-negativity constraint. The clamp operation enforces physical non-negativity of the predicted distances.

\paragraph{Gated Fusion module.}
Finally, the Gated Fusion module combines the residual-corrected prediction and the geometry-derived distance according to
\begin{equation}
pred =
\bigl(
g \odot d_{\mathrm{refined}}
+
(1-g) \odot l_d
\bigr)
\odot m_{\mathrm{seq}}
\end{equation}
This module is deterministic and contains no learnable parameters. Its purpose is to balance two complementary sources of information: the learned denoising estimate and the geometry-consistent distance implied by the latent layout. When the learned temporal correction is reliable, the gate can place more weight on $d_{\mathrm{refined}}$; when geometric consistency is more informative, the fusion can lean toward $l_d$. The final output is the denoised distance tensor $pred \in \mathbb{R}^{B \times T \times N}.$

\paragraph{Anchor reconstruction. }
To evaluate whether the predicted distances preserve geometric consistency, we reconstruct anchor layouts from the denoised distance predictions. Given the final output $pred \in \mathbb{R}^{B \times T \times N}$, we consider a single episode and denote the predicted distance between the vehicle and anchor $i$ at time $t$ as $\hat d_{t,i}$. Given the corresponding vehicle trajectory $\{p_t\}_{t=1}^{T}$, where $p_t \in \mathbb{R}^2$ denotes the 2D tag location at time step $t$, we estimate the anchor position $\hat a_i \in \mathbb{R}^2$ by solving a weighted nonlinear least-squares problem:
\begin{equation}
\hat a_i = \arg\min_{a \in \mathbb{R}^2} \sum_{t=1}^{T} \omega_{t,i} \left( \left\| p_t - a \right\|_2 - \hat d_{t,i} \right)^2,
\end{equation}
where $i$ indexes the anchor and $T$ is the number of time steps used for reconstruction.

The weight $\omega_{t,i}$ is defined based on propagation conditions:
\begin{equation}
\omega_{t,i} =
\begin{cases}
1, & \text{if LOS},\\
0.2, & \text{if NLOS}.
\end{cases}
\end{equation}

This reconstruction step provides a geometry-level interpretation of the predicted distances: if the distances are globally consistent, the recovered anchor layout should align with the true spatial configuration.

\paragraph{Framework summary.}
Overall, GAIA integrates temporal refinement, latent geometric layout inference, and geometry-consistent distance projection into a unified denoising framework for work-zone boundary reconstruction. The experimental results show that GAIA improves boundary-level reconstruction over the evaluated baselines in both real outdoor deployments and supplementary stress-test simulations. In particular, the framework improves boundary-level metrics such as IoU while remaining competitive on ranging accuracy (Table~\ref{tab:real_results} and Table~\ref{tab:overall_results}).

\subsection{Training Strategy}
\label{sec:training}

Algorithm~\ref{alg:gaia_training} summarizes the training procedure of GAIA.
During training, each mini-batch contains the vehicle location sequence
$p_{\mathrm{seq}} \in \mathbb{R}^{B \times T \times 2}$,
the observed UWB range sequence
$d_{\mathrm{seq}}^{\mathrm{obs}} \in \mathbb{R}^{B \times T \times N}$,
and the validity mask
$m_{\mathrm{seq}} \in \{0,1\}^{B \times T \times N}$.
In addition to these inputs, training uses the ground-truth distance sequence
$d_{\mathrm{seq}}^{\mathrm{gt}} \in \mathbb{R}^{B \times T \times N}$ as the primary supervision target.
When available, we further use an NLOS mask
$m_{\mathrm{seq}}^{\mathrm{NLOS}} \in \{0,1\}^{B \times T \times N}$,
which indicates whether a valid range observation is affected by NLOS propagation,
and the ground-truth anchor layout
$A^{\mathrm{gt}} \in \mathbb{R}^{B \times N \times 2}$,
which provides the true 2D anchor positions for geometry-related supervision.

\begin{algorithm}[H]
\caption{Training procedure for GAIA}
\label{alg:gaia_training}
\begin{algorithmic}[1]
\Require $p_{\mathrm{seq}}, d_{\mathrm{seq}}^{\mathrm{obs}}, m_{\mathrm{seq}}, d_{\mathrm{seq}}^{\mathrm{gt}}, m_{\mathrm{seq}}^{\mathrm{NLOS}}, A^{\mathrm{gt}}$
\Ensure Updated model parameters $\theta$

\For{each epoch}
    \For{each batch}
        \State $d_{\mathrm{use}}^{\mathrm{obs}} \leftarrow d_{\mathrm{seq}}^{\mathrm{obs}} \odot m_{\mathrm{seq}}$
        \State $(pred, l, l_d) \leftarrow \mathrm{GAIA}(p_{\mathrm{seq}}, d_{\mathrm{use}}^{\mathrm{obs}}, m_{\mathrm{seq}})$
        
        \State $w \leftarrow \mathrm{WeightMask}(m_{\mathrm{seq}}, m_{\mathrm{seq}}^{\mathrm{NLOS}})$
        
        \State $\mathcal{L}_{\mathrm{mask}} \leftarrow \mathrm{MaskedLoss}(pred, d_{\mathrm{seq}}^{\mathrm{gt}}, w)$
        \State $\mathcal{L}_{\mathrm{layout}}, \mathcal{L}_{\mathrm{cons}}, \mathcal{L}_{\mathrm{anchor}} \leftarrow 0,\, 0,\, 0$

        \If{$A^{\mathrm{gt}}$ available}
            \State $\mathcal{L}_{\mathrm{layout}}, \mathcal{L}_{\mathrm{cons}} \leftarrow \mathrm{LayoutLoss}(l, l_d, pred, A^{\mathrm{gt}})$
            \State $\mathcal{L}_{\mathrm{anchor}} \leftarrow \mathrm{AnchorLoss}(pred, p_{\mathrm{seq}}, m_{\mathrm{seq}}, m_{\mathrm{seq}}^{\mathrm{NLOS}}, A^{\mathrm{gt}})$
        \EndIf
        
        \State $\mathcal{L} \leftarrow \mathcal{L}_{\mathrm{mask}} 
        + \lambda_1 \mathcal{L}_{\mathrm{layout}} 
        + \lambda_2 \mathcal{L}_{\mathrm{cons}} 
        + \lambda_3 \mathcal{L}_{\mathrm{anchor}}$
        
        \State $\theta \leftarrow \theta - \eta \nabla_{\theta} \mathcal{L}$
    \EndFor
\EndFor
\end{algorithmic}
\end{algorithm}

For each batch, we first construct the masked observation
$d_{\mathrm{use}}^{\mathrm{obs}}$,
so that padded or invalid anchor measurements do not contribute to the forward pass.
The batch is then fed into GAIA, which produces three outputs:
the final denoised distance prediction
$pred \in \mathbb{R}^{B \times T \times N}$,
the latent anchor layout
$l \in \mathbb{R}^{B \times N \times 2}$,
and the geometry-derived distance tensor
$l_d \in \mathbb{R}^{B \times T \times N}$.
Here, $pred$ is the primary output used for distance supervision,
$l$ is the inferred 2D anchor configuration,
and $l_d$ is the deterministic distance field implied by the predicted layout and vehicle trajectory.

After the forward pass, we construct a supervision weight tensor
$w \in \mathbb{R}^{B \times T \times N}$ from the validity mask $m_{\mathrm{seq}}$ and, when available, the NLOS mask $m_{\mathrm{seq}}^{\mathrm{NLOS}}$.
This weighting step ensures that the loss is evaluated only on valid anchor observations and allows different emphasis to be assigned to LOS and NLOS samples.
Using this weight tensor, we compute the masked distance loss
$\mathcal{L}_{\mathrm{mask}}$ between the denoised prediction $pred$ and the ground-truth distance sequence $d_{\mathrm{seq}}^{\mathrm{gt}}$.
This is the primary supervision term and trains the model to recover accurate per-anchor denoised ranges on valid measurements.

When the ground-truth anchor layout $A^{\mathrm{gt}}$ is available, we additionally apply geometry-related supervision.
Specifically, the layout supervision branch produces two terms.
The first is a layout loss
$\mathcal{L}_{\mathrm{layout}}$,
which encourages the predicted latent layout $l$ to match the true anchor geometry $A^{\mathrm{gt}}$.
The second is a consistency term
$\mathcal{L}_{\mathrm{cons}}$,
which is produced together with the layout supervision and encourages agreement between the final predicted distances $pred$ and the geometry-derived distances $l_d$.
This consistency term couples denoising with the inferred spatial structure, discouraging the final prediction from drifting too far from the geometry implied by the latent layout.

In addition, we optionally apply an anchor-level geometry loss
$\mathcal{L}_{\mathrm{anchor}}$.
This term is computed by performing differentiable multilateration on a temporally subsampled sequence derived from the predicted distances $pred$, the vehicle trajectory $p_{\mathrm{seq}}$, and a weighted validity mask constructed from $m_{\mathrm{seq}}$ and the NLOS mask $m_{\mathrm{seq}}^{\mathrm{NLOS}}$; the recovered anchor positions are then compared against the ground-truth anchor layout $A^{\mathrm{gt}}$.
Compared with the direct layout supervision term, this loss acts as a downstream geometric constraint because it evaluates whether the predicted ranges are sufficient for recovering the correct anchor geometry through multilateration.

The total training objective is then formed as a weighted sum of all active loss terms:
the masked distance loss $\mathcal{L}_{\mathrm{mask}}$,
the layout supervision loss $\mathcal{L}_{\mathrm{layout}}$,
the consistency term $\mathcal{L}_{\mathrm{cons}}$,
and the anchor-level geometry loss $\mathcal{L}_{\mathrm{anchor}}$.
The corresponding balancing coefficients are denoted by
$\lambda_1$, $\lambda_2$, and $\lambda_3$ for the auxiliary geometric terms.
In this way, the primary denoising objective is preserved, while the additional supervision terms regularize the model toward geometry-consistent predictions.

Finally, the total loss is backpropagated to update the trainable parameters of GAIA.
The PoseMLP Base module remains fixed during this process because it is pretrained and frozen, while the remaining trainable modules are optimized end-to-end.
Through this procedure, GAIA combines masked range supervision, latent layout supervision, geometric consistency, and anchor-level reconstruction constraints within a unified training framework.

\subsection{Loss Functions}
\label{sec:loss}

The training objective of GAIA consists of four components:
a masked distance loss, a layout supervision loss, a geometry consistency loss,
and an anchor-level geometry loss. The final objective is a weighted sum of these terms.

\paragraph{Masked distance loss.}
The primary supervision term is a weighted masked Huber loss between the predicted distances $pred$ and the ground-truth distances $d_{\mathrm{seq}}^{\mathrm{gt}}$. 
We first define the residual as
\begin{equation}
r_{\mathrm{hub}} = pred - d_{\mathrm{seq}}^{\mathrm{gt}},
\end{equation}
where $r_{\mathrm{hub}} \in \mathbb{R}^{B \times T \times N}$ denotes the element-wise distance error, and $\delta_{\mathrm{hub}} > 0$ denotes the Huber threshold. The loss is defined as
\begin{equation}
\mathcal{L}_{\mathrm{mask}} =
\frac{\sum w \odot \ell_{\mathrm{Huber}}\!\left(r_{\mathrm{hub}}\right)}
{\sum w},
\quad
\ell_{\mathrm{Huber}}\!\left(r_{\mathrm{hub}}\right)=
\begin{cases}
\frac{1}{2} r_{\mathrm{hub}}^2, & |r_{\mathrm{hub}}| \le \delta_{\mathrm{hub}},\\[4pt]
\delta_{\mathrm{hub}} \left(|r_{\mathrm{hub}}| - \frac{1}{2}\delta_{\mathrm{hub}}\right), & |r_{\mathrm{hub}}| > \delta_{\mathrm{hub}}.
\end{cases}
\end{equation}

This loss serves as the primary denoising objective. The weighting tensor $w$ restricts supervision to valid observations and allows different emphasis on LOS and NLOS measurements, while the Huber formulation improves robustness to large residuals caused by burst noise or severe NLOS effects.

\paragraph{Layout supervision loss.}
To supervise the latent anchor layout, we define a regression loss between the predicted layout $l$ and the ground-truth anchor positions $A^{\mathrm{gt}}$:
\begin{equation}
\mathcal{L}_{\mathrm{layout}} =
\frac{1}{\sum m_{\mathrm{anchor}}}
\sum m_{\mathrm{anchor}} \cdot \left\| l - A^{\mathrm{gt}} \right\|^2,
\end{equation}
where the anchor-level validity mask marks anchors that are observed at least once in the sequence,
\begin{equation}
m_{\mathrm{anchor}} = \mathbb{I}\!\left(\max_{t} m_{\mathrm{seq}}[:,t,:] > 0.5\right),
\end{equation}
with $m_{\mathrm{anchor}} \in \{0,1\}^{B \times N}$. In our data the per-anchor validity is constant over a sequence, so this is equivalent to inspecting any single frame.

This loss encourages the model to infer a geometrically consistent anchor configuration from coupled temporal ranging observations.

\paragraph{Geometry consistency loss.}
In addition to direct layout supervision, we introduce a consistency term that aligns the final predicted distances with the geometry-derived distances:
\begin{equation}
\mathcal{L}_{\mathrm{cons}} =
\frac{1}{\sum m_{\mathrm{seq}}}
\sum m_{\mathrm{seq}} \cdot \left( pred - l_d \right)^2.
\end{equation}
This term is produced together with the layout supervision branch and encourages consistency between the learned denoising output and the distances implied by the inferred layout.
As a result, the model is encouraged to produce predictions that are accurate and geometrically coherent.

\paragraph{Anchor-level geometry loss.}
To further encourage geometric correctness, we introduce an anchor-level loss based on differentiable multilateration.
For computational efficiency, this loss is evaluated on a temporally subsampled sequence.
Let $pred_{\mathrm{s}}$, $p_{\mathrm{seq}}^{s}$, and $m_{\mathrm{seq}}^{s}$ denote the temporally subsampled predicted distances, vehicle trajectory, and validity mask, respectively.
From $m_{\mathrm{seq}}^{s}$ and the corresponding subsampled NLOS mask, we construct an anchor-level weighting tensor
$w_{\mathrm{anchor}} \in \mathbb{R}^{B \times T_s \times N}$,
which is used only in this multilateration branch.

We then recover anchor positions through an unrolled multilateration operator:
\begin{equation}
(\hat{A}, \mathcal{R}_{\mathrm{geo}}) = \mathrm{Multilateration}(pred_{\mathrm{s}}, p_{\mathrm{seq}}^{s}, w_{\mathrm{anchor}}),
\end{equation}
where $\hat{A} \in \mathbb{R}^{B \times N \times 2}$ denotes the recovered anchor positions and $\mathcal{R}_{\mathrm{geo}}$ is the residual of the geometric fitting process.
In practice, the multilateration procedure solves for each anchor independently from the predicted distance sequence and the vehicle trajectory, using a least-squares initialization followed by iterative Gauss--Newton refinement.
The residual term $\mathcal{R}_{\mathrm{geo}}$ measures the mismatch between the recovered geometry and the predicted distances after reconstruction.

To include only anchors valid in the subsampled sequence in the supervision, we define a subsampled anchor-validity mask
\begin{equation}
m_{\mathrm{anchor}}^{\mathrm{sub}} = \mathbb{I}\!\left(\max_{t} m_{\mathrm{seq}}^{s}[:,t,:] > 0.5\right),
\end{equation}
with $m_{\mathrm{anchor}}^{\mathrm{sub}} \in \{0,1\}^{B \times N}$.
The anchor-level geometry loss is then defined as
\begin{equation}
\mathcal{L}_{\mathrm{anchor}} =
\frac{1}{\sum m_{\mathrm{anchor}}^{\mathrm{sub}}}
\sum m_{\mathrm{anchor}}^{\mathrm{sub}} \cdot \left\| \hat{A} - A^{\mathrm{gt}} \right\|^2
+ \mathcal{R}_{\mathrm{geo}}.
\end{equation}

This loss evaluates whether the predicted distances are sufficient to recover the correct anchor geometry through multilateration, thereby providing a downstream geometric constraint that complements the layout supervision branch.

\paragraph{Total loss.}
The overall training objective is defined as
\begin{equation}
\mathcal{L} =
\mathcal{L}_{\mathrm{mask}}
+ \lambda_1 \mathcal{L}_{\mathrm{layout}}
+ \lambda_2 \mathcal{L}_{\mathrm{cons}}
+ \lambda_3 \mathcal{L}_{\mathrm{anchor}},
\end{equation}
where $\lambda_1$, $\lambda_2$, and $\lambda_3$ are weighting coefficients that balance the contribution of each term.
The masked distance loss acts as the primary supervision, while the remaining terms regularize the model toward geometry-consistent predictions.

\section{Experiments}

\subsection{Experiment Setup}

We evaluate GAIA primarily on a real-world outdoor UWB dataset containing synchronized UWB, GNSS, and IMU measurements under dynamic LOS and NLOS scenarios \cite{lee2025comprehensive}. This dataset serves as the main empirical basis of the study. It includes multi-anchor UWB ranging together with high-accuracy trajectory references, enabling evaluation of range denoising, anchor reconstruction, and downstream work-zone boundary reconstruction under practical outdoor sensing conditions.

In addition to the real-world evaluation, we use a supplementary stress-test simulator to examine model behavior under more severe ranging conditions. The simulator follows the V2I work-zone UWB reconstruction setting in \cite{liu2026v2iworkzonegeometry} and is calibrated using summary error statistics from the real dataset. It is used for controlled robustness analysis and ablation, not as a replacement for real-world validation.

Although the simulator is calibrated using real UWB measurements, it does not duplicate the real dataset. Calibration is applied only to the ranging-error statistics, while the simulated episodes provide controlled variations in anchor layouts, vehicle trajectories, NLOS severity, burst noise, and long-tail errors. This makes the simulator useful for stress testing and component-level analysis beyond the limited coverage of the real dataset.

We compare GAIA with Kalman~\cite{zhang2016kalman}, MLP~\cite{wymeersch2012machine}, PoseKalman~\cite{liu2026v2iworkzonegeometry}, and PoseMLP~\cite{liu2026v2iworkzonegeometry}. We evaluate performance using \texttt{distance\_mse\_overall}, \texttt{distance\_mse\_los}, and \texttt{distance\_mse\_nlos} for range prediction; anchor position error for anchor estimation; and \texttt{polygon\_iou} and \texttt{polygon\_hausdorff} for geometry reconstruction.

\subsection{Dataset}

Our main experiments use a real-world outdoor UWB dataset containing synchronized UWB, GNSS, and IMU measurements \cite{lee2025comprehensive}. The dataset provides multi-anchor UWB range observations, reference tag trajectories, anchor locations, and LOS/NLOS propagation conditions. These measurements allow us to evaluate signal-level range denoising as well as downstream anchor reconstruction and work-zone boundary reconstruction.

For the dynamic setting, each episode consists of a time-ordered vehicle trajectory 
\(\{(x_t, y_t)\}_{t=1}^{T}\), together with corresponding UWB observations 
\(\{d_{t,i}^{\mathrm{obs}}\}_{t=1}^{T}\) from multiple anchors. The position of the \(i\)-th anchor is denoted by 
\((x_i^{a}, y_i^{a})\), and the tag position at time step \(t\) is denoted by 
\((x_t, y_t)\). The observed UWB distance between the tag and anchor \(i\) at time step \(t\) is denoted by 
\(d_{t,i}^{\mathrm{obs}}\).

\begin{figure}[]
    \centering
    \includegraphics[width=\linewidth]{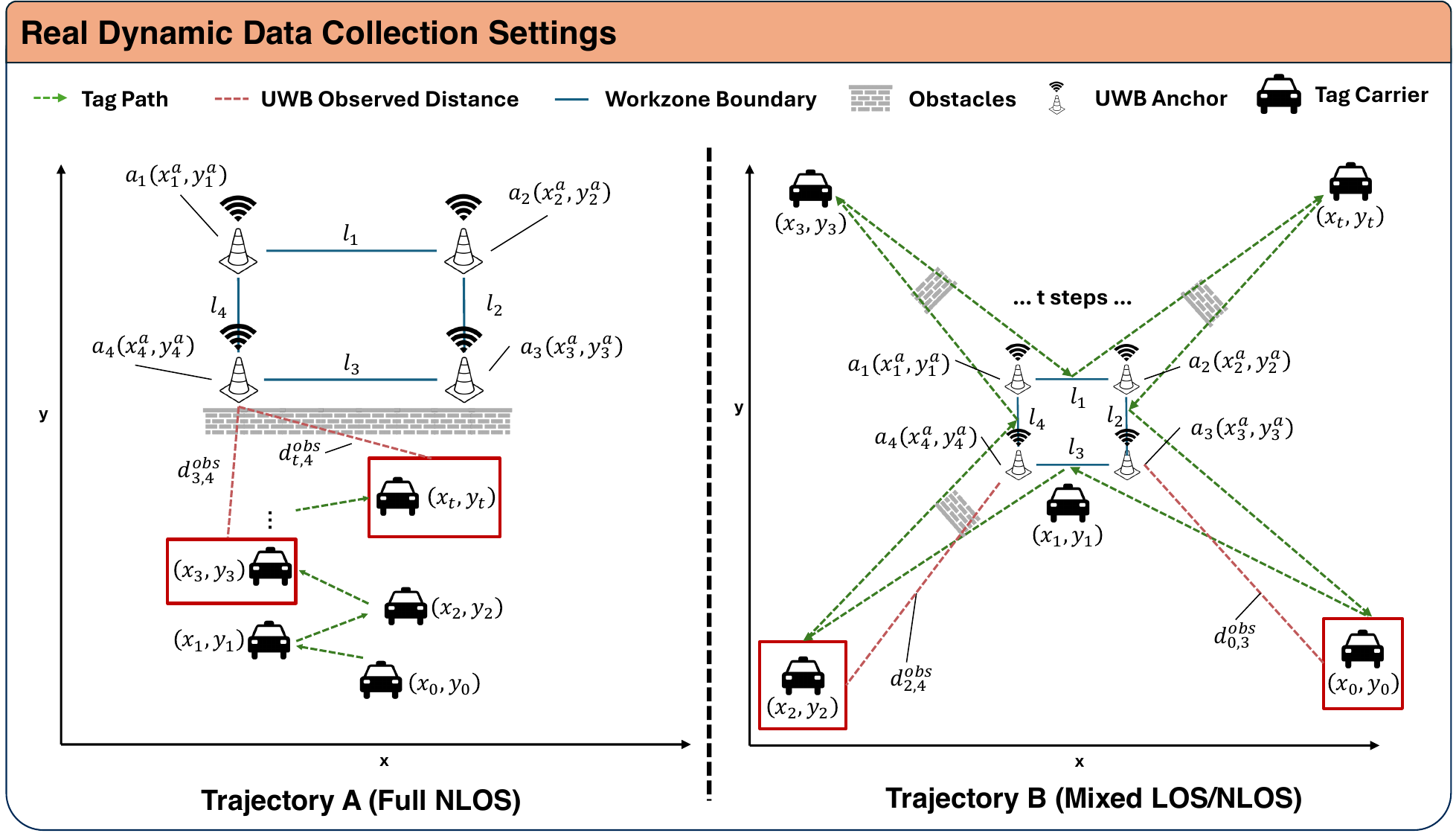}
    \caption{Real dynamic data collection settings. Trajectory A corresponds to a full-NLOS setting, while Trajectory B corresponds to a mixed LOS/NLOS setting. The figure is adapted and re-illustrated based on the outdoor UWB dataset in \cite{lee2025comprehensive}.}
    \label{fig:real_dynamic_settings}
\end{figure}

The dataset contains dynamic measurements under both full-NLOS and mixed LOS/NLOS conditions, as shown in Fig.~\ref{fig:real_dynamic_settings}. Trajectory A corresponds to a full-NLOS setting, where the tag moves behind obstacles with respect to the anchor array and most measurements are collected under blocked conditions. Trajectory B corresponds to a mixed LOS/NLOS setting, where the tag moves through regions with different visibility conditions relative to the anchors. These two trajectory patterns provide practical outdoor UWB measurements for evaluating geometry-aware denoising under different propagation conditions.

\subsection{Supplementary Stress-Test Simulation}
\label{sec:Simulation}

Because real-world UWB data are costly to collect and may not cover all severe NLOS and long-tail error cases, we include a supplementary stress-test simulation to examine model robustness under more challenging ranging conditions. This simulation serves as a controlled extension of the real-data evaluation.

The simulator generates multi-anchor UWB ranging sequences under controlled work-zone-like geometry. Following \cite{liu2026v2iworkzonegeometry}, it includes vehicle trajectories, anchor layouts, LOS/NLOS propagation regimes, temporally correlated errors, and occasional long-tail outliers. We calibrate the noise statistics using the real outdoor UWB dataset so that simulated ranging errors better match the empirical LOS/NLOS error distribution. The simulator serves as a controlled stress-test environment for evaluating whether geometry-aware denoising remains effective under stronger and more diverse ranging noise than the real dataset alone provides.

The simulated stress-test data are used for two purposes: first, to evaluate whether GAIA remains effective when UWB measurements contain stronger NLOS and long-tail errors than those observed in the main real-world dataset; and second, to support ablation studies that isolate the contributions of the Layout Head, GeoDist module, and frozen PoseMLP Base. The main deployment-oriented conclusions are based on the real-world UWB evaluation, while the simulation results provide supplementary robustness and component-level evidence.

\subsection{Baselines and Evaluation Metrics}
\paragraph{\textbf{Baseline Selection.}}
We compare GAIA with four baselines: Kalman~\cite{zhang2016kalman}, MLP~\cite{wymeersch2012machine}, PoseKalman~\cite{liu2026v2iworkzonegeometry}, and PoseMLP~\cite{liu2026v2iworkzonegeometry}. These baselines cover two modeling dimensions: filtering-based versus learning-based denoising, and pose-agnostic versus pose-aware input representations.

Kalman represents a classical filtering-based approach that operates on sequential UWB range observations. We include it as an interpretable sequential smoothing baseline, following prior use of Kalman filtering for UWB positioning under LOS/NLOS scenarios~\cite{zhang2016kalman}. MLP is a feedforward learning baseline without explicit temporal structure or pose input. It represents machine-learning-based UWB ranging-error mitigation methods that learn direct corrections from ranging observations~\cite{wymeersch2012machine}.

PoseMLP follows the pose-conditioned feedforward baseline used in prior V2I work-zone UWB geometry reconstruction~\cite{liu2026v2iworkzonegeometry}. It extends the MLP baseline by incorporating vehicle pose as an additional input. PoseKalman follows the pose-aware filtering baseline in the same V2I work-zone reconstruction protocol~\cite{liu2026v2iworkzonegeometry}, extending Kalman-style filtering with pose-related information. Together, these four baselines form a compact comparison space that spans filtering versus learning and pose-agnostic versus pose-aware settings, enabling a structured evaluation of denoising strategies under the same experimental protocol.

\paragraph{\textbf{Evaluation metrics.}}
For clarity, we describe all metrics at the level of a single episode.
We use scalar- and vector-level notation, where each quantity corresponds to a specific time step $t$ and anchor index $i$.

\textbf{Range-level metrics.}
Let $\hat d_{t,i}$ denote the denoised distance between the vehicle and anchor $i$ at time $t$, and let $d^{\mathrm{gt}}_{t,i}$ denote the corresponding ground-truth distance.
We use $m_{t,i} \in \{0,1\}$ to indicate whether the measurement is valid, and $n_{t,i} \in \{0,1\}$ to indicate whether the measurement is affected by NLOS.

\begin{equation}
\mathrm{MSE}_{\mathrm{overall}} =
\frac{\sum_{t,i} m_{t,i} \left(\hat d_{t,i} - d^{\mathrm{gt}}_{t,i}\right)^2}
{\sum_{t,i} m_{t,i}}.
\end{equation}

\begin{equation}
\mathrm{MSE}_{\mathrm{LOS}} =
\frac{\sum_{t,i} m_{t,i}(1-n_{t,i}) \left(\hat d_{t,i} - d^{\mathrm{gt}}_{t,i}\right)^2}
{\sum_{t,i} m_{t,i}(1-n_{t,i})},
\end{equation}

\begin{equation}
\mathrm{MSE}_{\mathrm{NLOS}} =
\frac{\sum_{t,i} m_{t,i} n_{t,i} \left(\hat d_{t,i} - d^{\mathrm{gt}}_{t,i}\right)^2}
{\sum_{t,i} m_{t,i} n_{t,i}}.
\end{equation}

\textbf{Anchor-level metrics.}
Let $a_i^{\mathrm{gt}} \in \mathbb{R}^2$ denote the ground-truth position of anchor $i$, which is constant over time.
Given the denoised distances $\hat d_{t,i}$ and the known trajectory $p_t^{\mathrm{gt}}$, we estimate a single anchor position $\hat a_i$ by solving a nonlinear least-squares problem over the entire episode. We report the mean anchor position error, defined as the average Euclidean distance between the estimated and ground-truth anchor positions:

\begin{equation}
\mathrm{MAE}_{\mathrm{anchor}} =
\frac{1}{|\mathcal{I}|}
\sum_{i \in \mathcal{I}} \left\| \hat a_i - a_i^{\mathrm{gt}} \right\|_2,
\end{equation}

where $\hat a_i$ is estimated using all time steps $\{t\}$ within the episode and $\mathcal{I} = \left\{ i \mid \sum_{t} m_{t,i} > 0 \right\}$ denotes the set of anchors that are observed at least once within the episode.

\textbf{Geometry-level metrics.}
To evaluate the reconstructed work-zone geometry, we construct polygons from anchor layouts. Specifically, the ground-truth polygon $\mathcal{S}^{\mathrm{gt}}$ is defined as the convex hull of $\{ a_i^{\mathrm{gt}} \}_{i \in \mathcal{I}}$, and the predicted polygon $\mathcal{S}^{\mathrm{pred}}$ is defined as the convex hull of $\{ \hat a_i \}_{i \in \mathcal{I}}$, where both $a_i^{\mathrm{gt}}$ and $\hat a_i$ represent episode-level anchor positions.

\begin{equation}
\mathrm{IoU} =
\frac{\mathrm{Area}(\mathcal{S}^{\mathrm{gt}} \cap \mathcal{S}^{\mathrm{pred}})}
{\mathrm{Area}(\mathcal{S}^{\mathrm{gt}} \cup \mathcal{S}^{\mathrm{pred}})}.
\end{equation}

\begin{equation}
d_H =
\max \left\{
\sup_{x \in \partial \mathcal{S}^{\mathrm{gt}}} \inf_{y \in \partial \mathcal{S}^{\mathrm{pred}}} \|x-y\|_2,\;
\sup_{y \in \partial \mathcal{S}^{\mathrm{pred}}} \inf_{x \in \partial \mathcal{S}^{\mathrm{gt}}} \|x-y\|_2
\right\}.
\end{equation}

The Hausdorff distance measures the worst-case boundary deviation between the two polygons, capturing extreme geometric errors that are not reflected by average metrics.

Together, these metrics form a hierarchical evaluation pipeline from signal-level accuracy to geometric reconstruction quality, following the progression from range denoising to anchor mapping and finally work-zone boundary reconstruction.

\section{Results}
\label{sec:Results}

\subsection{Main Results on the Real UWB Dataset}

We first evaluate GAIA on the real-world outdoor UWB dataset, which serves as the main empirical basis of this study. Table~\ref{tab:real_results} summarizes the results.

On the real-world dataset, GAIA achieves the lowest range MSE and the highest polygon IoU among the evaluated methods. It attains an overall distance MSE of 0.1414 and a polygon IoU of 0.2390. Compared with PoseMLP, the strongest learning-based baseline in this setting, GAIA reduces overall MSE by 18.4\% and increases polygon IoU by 15.5\%. GAIA also obtains the lowest LOS and NLOS MSE, indicating that the geometry-aware design improves range denoising under both propagation regimes.

GAIA's advantage is most evident in boundary overlap. While PoseMLP achieves a slightly lower anchor position error (2.1703 vs.\ 2.2493) and a lower Hausdorff distance (3.2905 vs.\ 3.7241), GAIA produces the highest polygon IoU, suggesting better overall boundary agreement. This indicates that geometry-aware denoising improves the global reconstructed work-zone shape, although worst-case boundary deviation remains an area for future improvement.

These results support the central motivation of this work: reliable work-zone reconstruction requires geometry-aware modeling beyond signal-level range accuracy. Geometry-aware modeling can improve the consistency among denoised ranges, reconstructed anchor layouts, and the resulting work-zone boundary.

\begin{table}[]
\centering
\caption{Results on the real-world UWB dataset. Baseline references: Kalman~\cite{zhang2016kalman}, MLP~\cite{wymeersch2012machine}, and PoseKalman/PoseMLP~\cite{liu2026v2iworkzonegeometry}. All methods are evaluated under the same protocol and use identical downstream solvers. Lower is better for all metrics except IoU.}
\label{tab:real_results}
\begin{tabular}{lcccccc}
\toprule
Method
& MSE$_{\text{overall}}$
& MSE$_{\text{LOS}}$
& MSE$_{\text{NLOS}}$
& MAE$_{\text{anchor}}$
& IoU
& Hausdorff \\
\midrule
Kalman & 0.1939 & 0.2385 & 0.1777 & 3.3650 & 0.0846 & 4.6569 \\
MLP & 0.5812 & 1.0111 & 0.4248 & 2.8809 & 0.0457 & 3.3946 \\
PoseKalman & 0.4733 & 0.1902 & 0.5762 & 12.9340 & 0.0159 & 15.8975 \\
PoseMLP & 0.1733 & 0.1102 & 0.1962 & \textbf{2.1703} & 0.2069 & \textbf{3.2905} \\
\textbf{GAIA} & \textbf{0.1414} & \textbf{0.1098} & \textbf{0.1529} & 2.2493 & \textbf{0.2390} & 3.7241 \\
\bottomrule
\end{tabular}
\end{table}

\subsection{Per-Episode Distribution Analysis}
\label{sec:perepisode}

Aggregate means can hide how reconstruction quality is distributed across episodes. We therefore analyze the per-episode distribution of the boundary-level metrics and report uncertainty using $95\%$ bootstrap confidence intervals over episodes ($10^3$ resamples). On the real dataset ($N=15$ episodes), GAIA attains a polygon IoU of $0.239$ $[0.157, 0.320]$ and a Hausdorff distance of $3.72$ $[2.15, 5.50]$; on the larger stress-test simulation ($N=216$), the intervals are tighter, with IoU $0.429$ $[0.383, 0.473]$ and Hausdorff $30.8$ $[27.2, 34.6]$. These boundary-metric means are consistent with Tables~\ref{tab:real_results} and~\ref{tab:overall_results}; the per-episode bootstrap quantifies their dispersion.

\begin{figure}[]
    \centering
    \includegraphics[width=\linewidth]{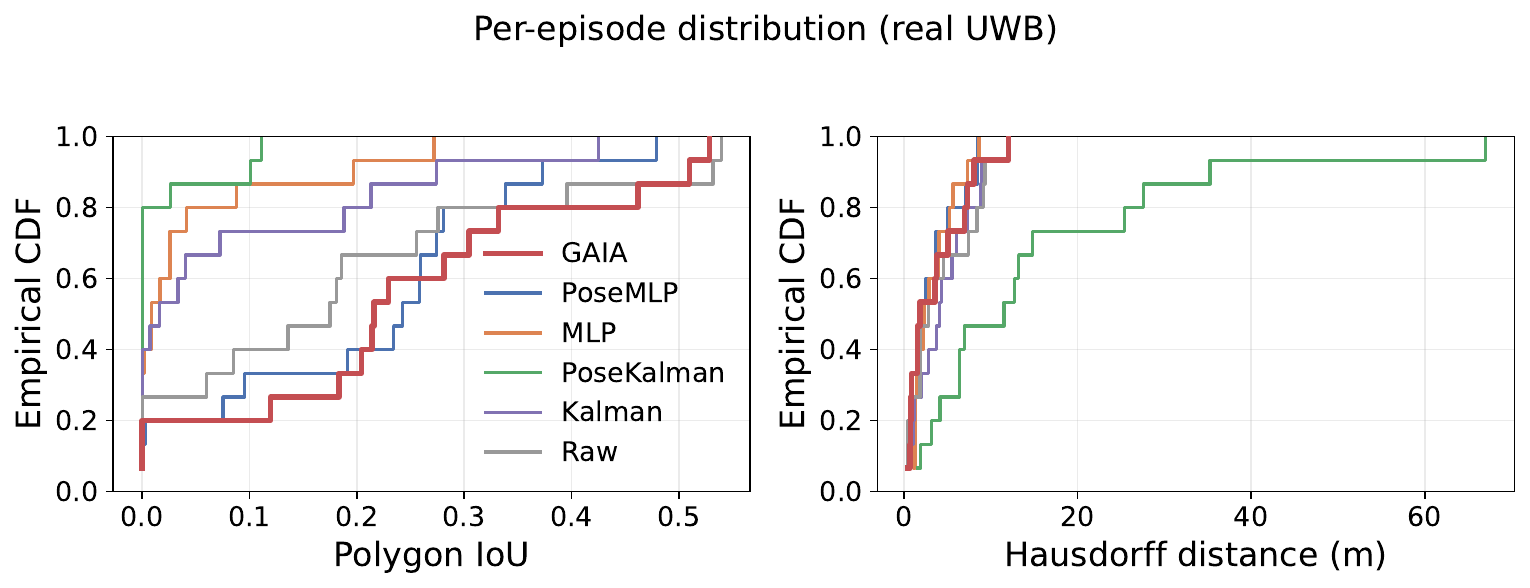}
    \caption{Per-episode empirical CDFs of polygon IoU (left; higher is better) and Hausdorff distance (right; lower is better) on the real UWB dataset. GAIA shows consistently strong IoU across episodes.}
    \label{fig:perep_cdf}
\end{figure}

\begin{figure}[]
    \centering
    \includegraphics[width=\linewidth]{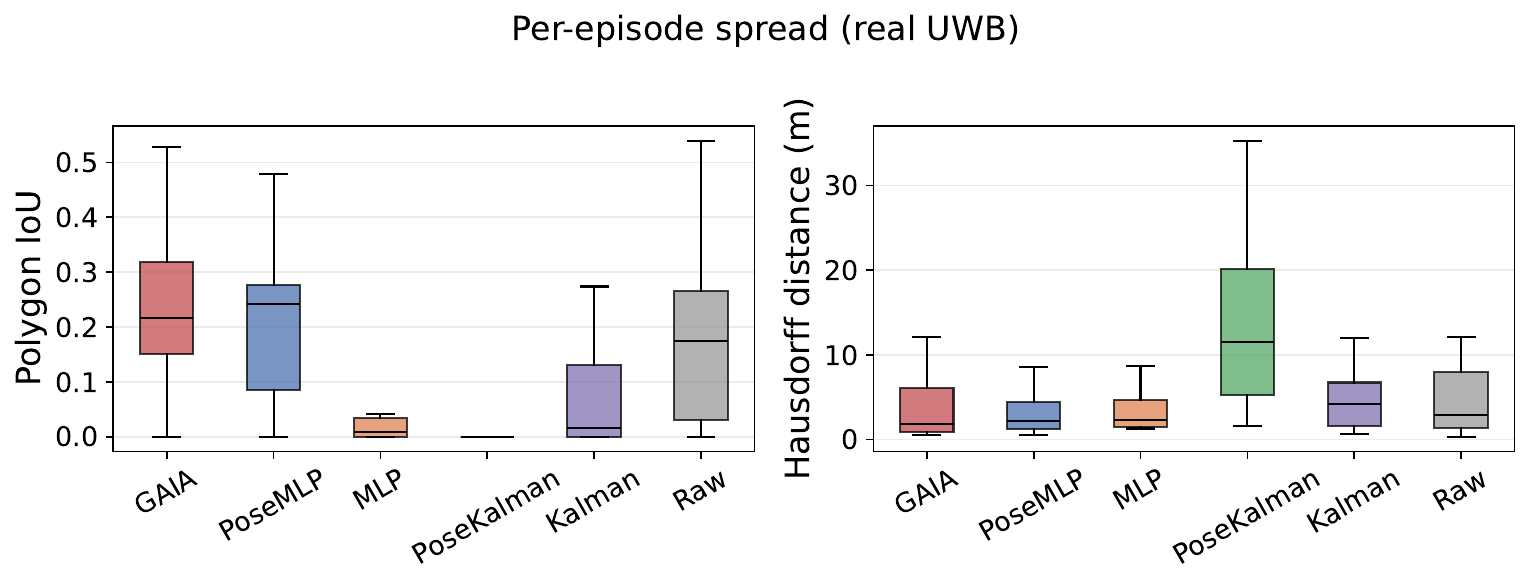}
    \caption{Per-episode spread of polygon IoU and Hausdorff distance on the real UWB dataset. GAIA shows a higher median IoU and a concentrated distribution across episodes.}
    \label{fig:perep_box}
\end{figure}

Figure~\ref{fig:perep_cdf} shows the empirical cumulative distribution functions (CDFs) of polygon IoU and Hausdorff distance. GAIA's IoU CDF lies to the right of the baselines across most of the range, indicating that its IoU advantage is not driven only by a few easy episodes. The corresponding box plots in Figure~\ref{fig:perep_box} show that GAIA also exhibits a higher median IoU and a concentrated spread. These distributional results indicate that the improvement in boundary overlap is consistent across episodes, while the Hausdorff results should be interpreted together with the mean values in Table~\ref{tab:real_results}.

\subsection{Supplementary Stress-Test Simulation Results}

We next evaluate all methods on the supplementary stress-test simulation. This setting exposes models to stronger NLOS corruption, long-tail ranging errors, and more diverse geometry variation than those covered by the real dataset. The purpose is to examine model robustness under controlled severe conditions while keeping real-world validation as the main empirical basis. The simulation setting follows \cite{liu2026v2iworkzonegeometry}, as described in Sec.~\ref{sec:Simulation}.

Table~\ref{tab:overall_results} summarizes the results. GAIA achieves the best performance across all reported metrics. Compared with PoseMLP, GAIA reduces the overall distance MSE by 50.8\%, lowers the anchor position error from 25.6580 to 19.7536, and increases polygon IoU from 0.3531 to 0.4286. Compared with PoseKalman, GAIA reduces overall MSE by 77.9\% and improves polygon IoU from 0.2519 to 0.4286.

These supplementary results are consistent with the real-world evaluation: explicitly modeling latent anchor layout and geometry-consistent distance projection improves downstream boundary reconstruction. The larger gains observed in simulation suggest that geometry-aware modeling is particularly beneficial under severe NLOS and long-tail ranging conditions.

\begin{table}[t]
\centering
\caption{Overall results on the supplementary stress-test simulation. Baseline references: Kalman~\cite{zhang2016kalman}, MLP~\cite{wymeersch2012machine}, and PoseKalman/PoseMLP~\cite{liu2026v2iworkzonegeometry}. Lower is better for all metrics except IoU.}
\label{tab:overall_results}
\begin{tabular}{lcccccc}
\toprule
Method
& MSE$_{\text{overall}}$
& MSE$_{\text{LOS}}$
& MSE$_{\text{NLOS}}$
& MAE$_{\text{anchor}}$
& IoU
& Hausdorff \\
\midrule
Kalman & 3.9913 & 1.4998 & 4.6750 & 32.3894 & 0.2404 & 62.5128 \\
MLP & 1.6514 & 1.2088 & 1.7728 & 30.7023 & 0.2668 & 54.4222 \\
PoseKalman & 1.7323 & 0.4998 & 2.0706 & 32.5667 & 0.2519 & 56.9666 \\
PoseMLP & 0.7773 & 0.6494 & 0.8124 & 25.6580 & 0.3531 & 37.8489 \\
\textbf{GAIA} & \textbf{0.3821} & \textbf{0.3567} & \textbf{0.3890} & \textbf{19.7536} & \textbf{0.4286} & \textbf{30.9233} \\
\bottomrule
\end{tabular}
\end{table}

\subsection{Robustness to Stressors}
\label{sec:robustness}

To probe how reconstruction degrades under harsher conditions, we perform a zero-shot robustness study. Starting from the held-out simulation test set, we apply a controlled stressor of increasing intensity to the observations, hold the work-zone geometry and trajectories fixed, and evaluate the same models without retraining. Holding the geometry fixed isolates the effect of each stressor and avoids confounding it with a change in spatial configuration. We study four stressors: additive range noise, outlier (heavy-tail) rate, amplification of the NLOS bias, and the number of available anchors.

\begin{figure}[]
    \centering
    \includegraphics[width=\linewidth]{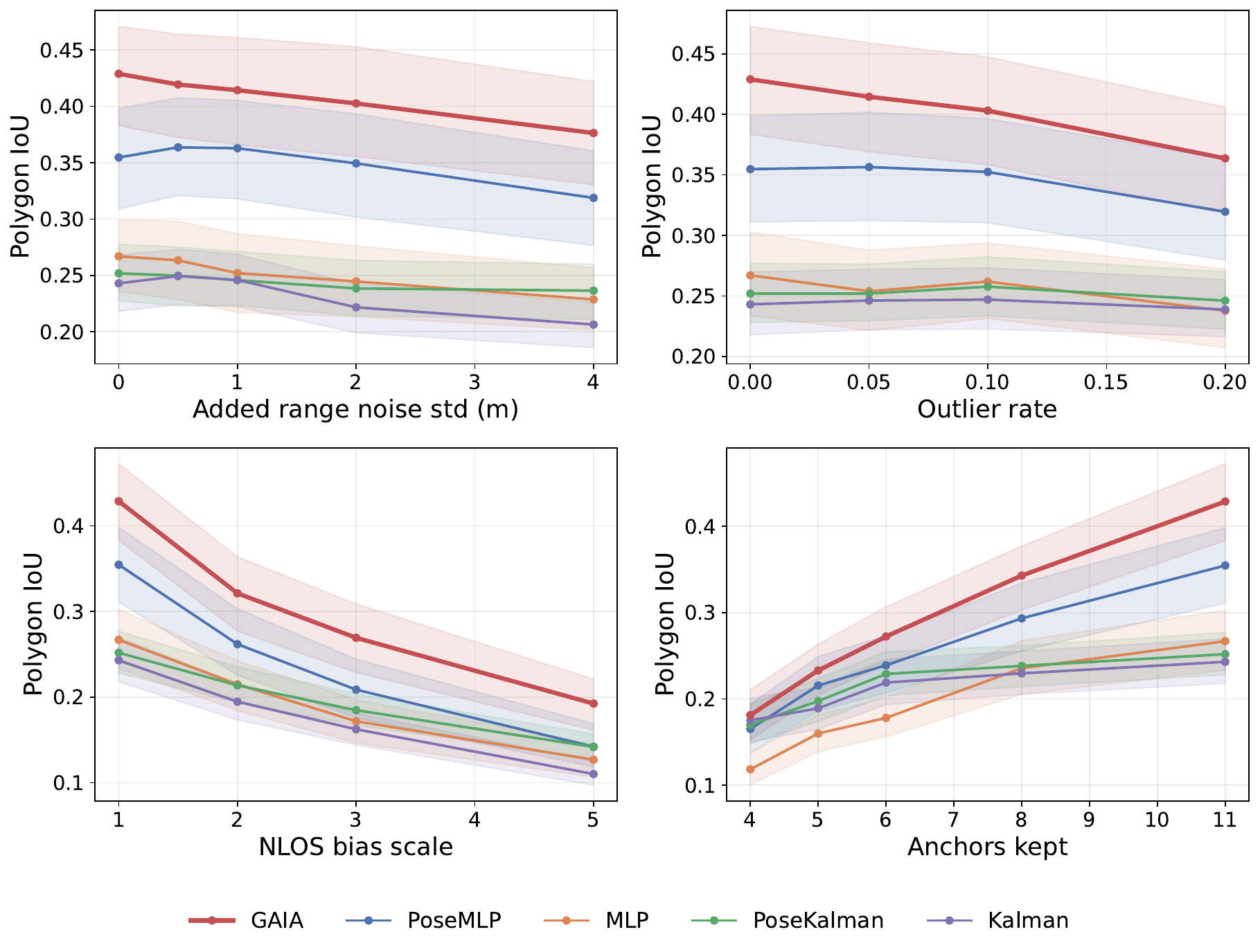}
    \caption{Zero-shot robustness of polygon IoU to four observation stressors (added range noise, outlier rate, NLOS-bias amplification, and anchor count), with $95\%$ bootstrap confidence bands. The same models are evaluated on the simulation test set with the stressor applied; geometry is held fixed. GAIA (red) attains the highest mean IoU across the range of every stressor, with the methods near-tied only at the most severe four-anchor setting.}
    \label{fig:robustness}
\end{figure}

Figure~\ref{fig:robustness} reports polygon IoU as a function of each stressor, with $95\%$ bootstrap confidence bands. GAIA attains the highest mean IoU across the full range of every stressor and degrades gradually: under $4$\,m of added range noise, its IoU falls from $0.429$ to $0.376$, and under a $4\times$ outlier rate, from $0.429$ to $0.364$. The most damaging stressors are amplifying the NLOS bias ($5\times$), which lowers IoU to $0.192$, and reducing the anchor count, where all methods converge as geometry becomes underdetermined. Table~\ref{tab:robustness_summary} summarizes GAIA's IoU at the base and most severe setting of each stressor, together with its mean margin over the best baseline at the most severe setting. GAIA retains a positive mean margin in every case; the margin remains clear under range noise, outliers, and NLOS-bias amplification, but narrows to near zero with only four anchors, where the geometry becomes too sparse for any method to reconstruct the boundary reliably and the methods are effectively tied within confidence intervals.

\begin{table}[t]
\centering
\caption{Robustness summary: GAIA polygon IoU at the base and most severe setting of each stressor, and GAIA's IoU margin over the best baseline at the most severe setting (positive means GAIA still leads).}
\label{tab:robustness_summary}
\begin{tabular}{lccc}
\toprule
Stressor (base $\rightarrow$ severe) & GAIA IoU (base) & GAIA IoU (severe) & Margin (severe) \\
\midrule
Range noise ($0 \rightarrow 4$\,m)        & 0.429 & 0.376 & $+0.058$ \\
Outlier rate ($0 \rightarrow 0.20$)        & 0.429 & 0.364 & $+0.044$ \\
NLOS bias scale ($1 \rightarrow 5\times$)  & 0.429 & 0.192 & $+0.050$ \\
Anchors kept (all $\rightarrow 4$)         & 0.429 & 0.181 & $+0.006$ \\
\bottomrule
\end{tabular}
\end{table}

\subsection{Ablation Study}
\label{sec:Ablation}

We conduct ablation experiments on the supplementary stress-test simulation to analyze the contribution of key GAIA components under controlled severe ranging conditions. Specifically, we consider three variants: (1) removing the Layout Head module, (2) removing the GeoDist module, and (3) unfreezing the PoseMLP Base and training it jointly with the rest of the model.

Table~\ref{tab:ablation} summarizes the results. Among all ablated variants, the strongest non-full model is the variant without geometry, which achieves an overall MSE of 0.5520. This indicates that even without explicit geometry modeling, the temporal refinement and learned components can still provide reasonable signal-level denoising.

However, a clear gap emerges in the downstream geometric metrics. Removing the Layout Head or GeoDist module degrades polygon IoU and Hausdorff distance, indicating that without explicit layout estimation and geometry-consistent distance modeling, the network struggles to recover coherent work-zone structures.

Among the ablations, the variant with the PoseMLP Base unfrozen achieves the highest polygon IoU (0.3799). Importantly, this variant still retains both the Layout Head and GeoDist modules, suggesting that the geometry-aware components are the main source of the reconstruction improvement.

Overall, these results show that the Layout Head and GeoDist modules are important for geometry reconstruction, with the PoseMLP Base serving mainly as a stable initialization.

\begin{table}[t]
\centering
\caption{Ablation study on the supplementary stress-test simulation. 
``w/o Layout Head'' removes the layout prediction module; 
``w/o GeoDist'' removes the geometry-based distance computation; 
``Unfreeze PoseMLP Base'' jointly trains the PoseMLP base network~\cite{liu2026v2iworkzonegeometry} as a trainable component. Lower is better for all metrics except IoU.}
\label{tab:ablation}
\begin{tabular}{lcccccc}
\toprule
Variant
& MSE$_{\text{overall}}$
& MSE$_{\text{LOS}}$
& MSE$_{\text{NLOS}}$
& MAE$_{\text{anchor}}$
& IoU
& Hausdorff \\
\midrule
w/o Layout Head & 0.6520 & 0.5817 & 0.6713 & 26.1689 & 0.3475 & 39.6997 \\
w/o GeoDist (w/o geometry) & 0.5520 & 0.4866 & 0.5699 & 24.9683 & 0.3654 & 36.1815 \\
Unfreeze PoseMLP Base & 0.7720 & 0.8036 & 0.7633 & 23.1187 & 0.3799 & 36.3602 \\
\textbf{GAIA} & 0.3821 & 0.3567 & 0.3890 & 19.7536 & 0.4286 & 30.9233 \\
\bottomrule
\end{tabular}
\end{table}

\subsection{Simulator Statistical Check}

We further report a statistical check of the supplementary simulator in Table~\ref{tab:sim_real_matching}. This comparison verifies that calibration moves the simulated ranging-error distribution closer to the measured outdoor UWB data than a naive simulator. The calibrated simulator better approximates the real tail behavior and outlier rate, although temporal burst persistence remains shorter than in the real measurements. Therefore, we use the simulator only for supplementary stress testing and ablation, while relying on real-data supervision for the main model evaluation.

\begin{table}[]
\centering
\caption{Statistical check of real-data-calibrated stress-test simulation against real ranging-error statistics.}
\label{tab:sim_real_matching}
\begin{tabular}{lcccccc}
\toprule
Source 
& Mean (m) 
& p95 (m) 
& p99 (m) 
& Outlier Rate 
& Burst Mean 
& Burst p95 \\
\midrule
Real Dataset 
& -0.0385
& 0.5127 
& 1.1279 
& 0.0120 
& 5.6889 
& 29.0 \\

Naive Simulator 
& 0.1680 
& 1.0564 
& 1.8520 
& 0.0563 
& 2.6068 
& 9.0 \\

Calibrated Simulator 
& 0.0662 
& 0.6303 
& 0.9009 
& 0.0066 
& 3.1784 
& 12.0 \\

\bottomrule
\end{tabular}
\end{table}

\begin{figure}[]
    \centering
    \includegraphics[width=0.6\linewidth]{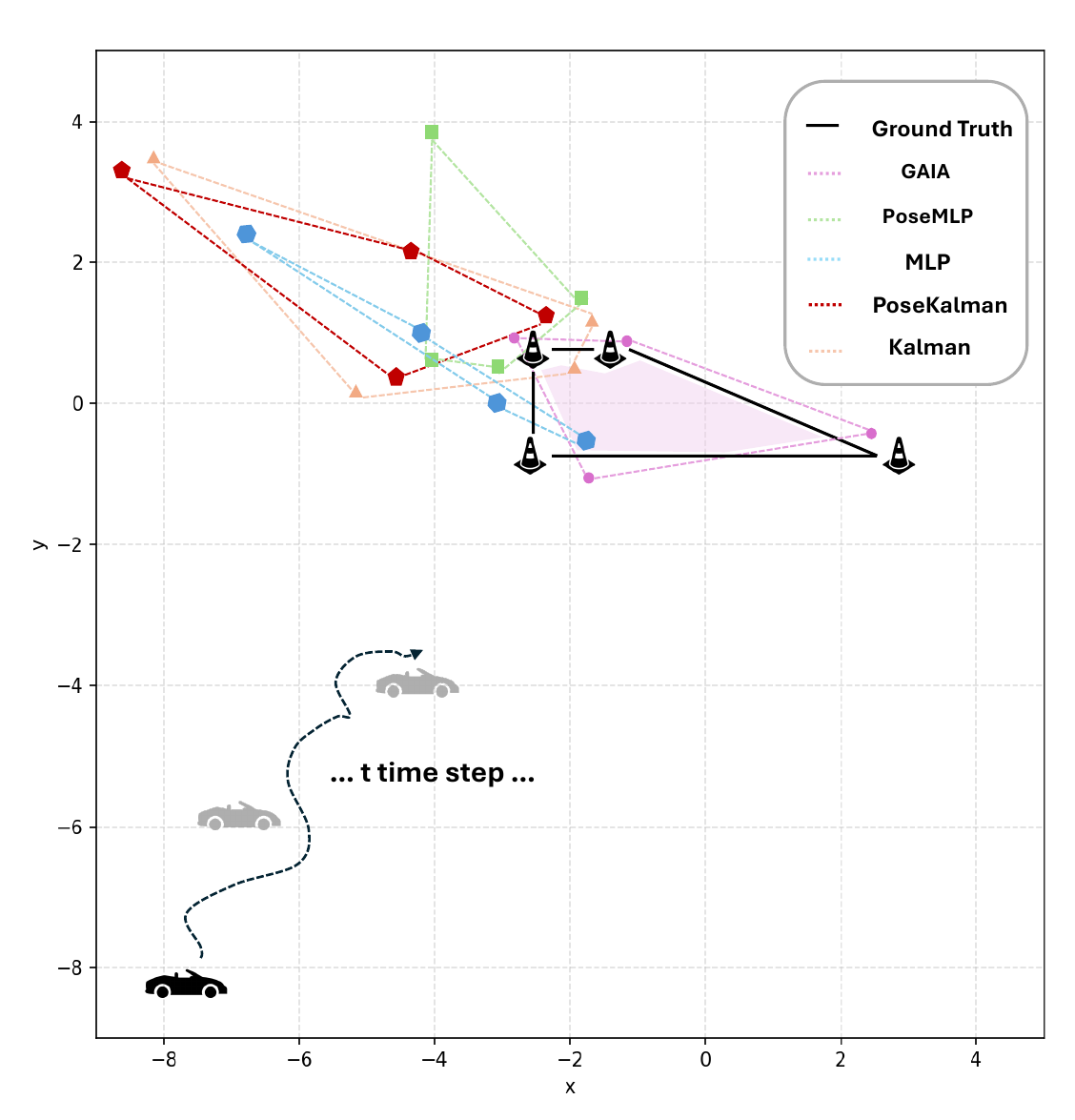}
    \caption{Qualitative comparison of anchor layout reconstruction. For each episode, anchor positions are reconstructed from predicted distances by solving a weighted nonlinear least-squares problem over a window of $T=64$ time steps, where LOS measurements are assigned weight 1 and NLOS measurements are down-weighted to 0.2. Ground-truth anchor positions are shown together with reconstructed layouts from different methods. The semi-transparent purple regions highlight the overlap between GAIA and the ground-truth geometry, which is substantial and closely follows the true work-zone shape. In contrast, baseline methods exhibit significant geometric distortion with minimal overlap.}
    \label{fig:qualitative_analysis}
\end{figure}

\subsection{Qualitative Analysis}

To further evaluate geometric consistency, we visualize anchor layouts reconstructed from the predicted distances using the formulation described in Sec.~\ref{sec:Architecture}. For each episode, the sequence is partitioned into segments of length $T=64$, and anchor positions are reconstructed accordingly. Fig.~\ref{fig:qualitative_analysis} shows a representative example. GAIA reconstructs anchor layouts that are closely aligned with the ground-truth work-zone geometry, with high spatial consistency across anchors.

In contrast, the baseline methods produce distorted layouts in which reconstructed anchors deviate substantially from their true positions, resulting in poor geometric alignment and large structural errors. These results show that GAIA preserves global geometric relationships beyond local range accuracy, supporting recovery of the underlying work-zone structure.

\section{Conclusion}

In this work, we propose GAIA, a geometry-aware framework for UWB range denoising and work-zone boundary reconstruction. GAIA jointly models temporal UWB ranging dynamics and latent anchor layout, and uses geometry-consistent distance projection to connect range correction with downstream spatial reconstruction. By incorporating spatial structure into the denoising process, GAIA produces distance predictions that are accurate at the signal level and more consistent with the reconstructed work-zone geometry.

Our evaluation is centered on a real-world outdoor UWB dataset with synchronized UWB, GNSS, and IMU measurements under LOS and NLOS conditions. On this dataset, GAIA achieves the lowest overall range MSE and the highest polygon IoU among the evaluated filtering-based and learning-based baselines. These results show that geometry-aware denoising improves boundary-level reconstruction beyond what can be achieved by optimizing range error alone. Supplementary real-data-calibrated stress-test simulation further supports the robustness of GAIA under controlled severe ranging conditions and provides a useful environment for ablation analysis. The main empirical conclusions are drawn from real outdoor UWB measurements, with simulation providing supplementary robustness evidence.

\bibliographystyle{cas-model2-names}

\bibliography{citations}

\end{document}